\documentclass[conference]{IEEEtran}
\IEEEoverridecommandlockouts
\usepackage{cite}
\usepackage[numbers]{natbib}

\usepackage{amsmath,amssymb,amsfonts}
\usepackage{algorithmic}
\usepackage{graphicx}
\usepackage{textcomp}
\usepackage{xcolor}
\usepackage{url}
\usepackage{changes}

\definechangesauthor[name={Glen}, color=orange]{CGH}

\def\BibTeX{{\rm B\kern-.05em{\sc i\kern-.025em b}\kern-.08em
    T\kern-.1667em\lower.7ex\hbox{E}\kern-.125emX}}
\begin{document}

\title{Collision-Free Inverse Kinematics Through QP Optimization (iKinQP)\\

}

\author{\IEEEauthorblockN{Julia Ashkanazy}
\IEEEauthorblockA{\textit{Naval Research Laboratory}}
\and
\IEEEauthorblockN{Ariana Spalter}
\IEEEauthorblockA{\textit{Naval Research Laboratory}}
\and
\IEEEauthorblockN{Joe Hays}
\IEEEauthorblockA{\textit{Naval Research Laboratory}}
\and
\IEEEauthorblockN{Laura Hiatt}
\IEEEauthorblockA{\textit{Naval Research Laboratory}}
\and
\IEEEauthorblockN{Roxana Leontie}
\IEEEauthorblockA{\textit{Naval Research Laboratory}}
\and
\IEEEauthorblockN{C. Glen Henshaw}
\IEEEauthorblockA{\textit{Naval Research Laboratory}}

% Washington, DC \\
}

\maketitle

\begin{abstract}
Robotic manipulators are often designed with more actuated degrees-of-freedom 
than required to fully control an end effector's position and orientation. These ``redundant'' 
manipulators can allow infinite joint configurations that satisfy a particular task-space position 
and orientation, providing more possibilities for the manipulator to traverse a smooth collision-free 
trajectory. However, finding such a trajectory is non-trivial because the inverse kinematics for redundant manipulators cannot typically be solved analytically. Many strategies have been developed to tackle this problem, including Jacobian pseudo-inverse method, rapidly-expanding-random tree (RRT) motion planning, and quadratic programming (QP) based methods. Here, we present a flexible inverse kinematics-based QP strategy (iKinQP). Because it is independent of robot dynamics, the algorithm is relatively light-weight, and able to run in real-time in step with torque control. Collisions are defined as kinematic trees of elementary geometries, making the algorithm  agnostic to the method used to determine what collisions are in the environment. Collisions are treated as hard constraints which guarantees the generation of collision-free 
trajectories. Trajectory smoothness is accomplished through the QP optimization. Our algorithm 
was evaluated for computational efficiency, smoothness, and its ability to provide trackable 
trajectories. It was shown that iKinQP is capable of providing smooth, collision-free trajectories 
at real-time rates.

\end{abstract}


\section{Introduction}

Redundant manipulators have more degrees-of-freedom (DOF) than necessary to fully define the end 
effector's position and orientation in space. The additional degrees-of-freedom extend the 
workspace of the robot, allowing for greater ease and flexibility in operating in cluttered, 
complex environments including manufacturing, space, disaster relief, and medical applications.

However, the additional DOFs also cause the redundancy resolution problem as multiple solutions to the inverse kinematics problem exist for redundant manipulators. This implies that multiple possible trajectories can guide the end effector from a start to end pose. This difficulty is exacerbated when extended to the case of multiple redundant manipulators operating within the same cluttered space. To solve such problems, constraints are typically added to resolve the redundancies, and ensure safety of the system and surrounding environment. For system safety, the trajectory must be smooth enough for the manipulator to follow without violating the manipulator control bandwidth and/or inducing any flexible modes, must avoid causing excessive torques or performing unnecessarily large motions, and must avoid collisions between manipulators or the environment. To prevent collisions with a constantly changing environment, and to compensate for trajectory tracking errors, trajectories cannot be precomputed, but rather must be computed on the fly. 

Our approach is based on a combination of Inverse Kinematics (iKin) and Quadratic Programming (QP), which we call iKinQP. We focus on solving the inverse kinematics for redundant manipulators where the extra actuated DOF leads to infinite solutions of the iKin problem. To do so, we utilize the numerical QP solver qpOASES \cite{Ferreau2014} to solve for our constrained optimization problem.

The system dynamics are simplified away in our approach by treating the mechanism as a 
first order linear dynamic system. It is assumed that the underlying controller designed to 
track the iKinQP generated trajectories compensates for the nonlinearities. The iKinQP 
algorithm easily allows a straightforward addition of dynamics, for instance catching a ball 
in flight or using a tool with significant uncontrolled dynamics.

Our approach was evaluated using a MuJoCo \cite{todorov2012mujoco} simulation of a Kinova 
Gen3 arm\footnote{\url{https://github.com/Kinovarobotics/ros_kortex/tree/noetic-devel/kortex_description/arms/gen3/7dof/urdf/GEN3_URDF_V12.urdf}}. 
Performance was assessed and compared for a single arm avoiding two different types of stationary environmental collisions, and for two arms dynamically avoiding each other. For each scenario, the performance was compared between an aggressive 5 second waypoint cadence, and a more relaxed 15 second waypoint cadence. During all tests waypoints were randomly selected along the surface of a 0.9 m radius sphere, and a spline was generated between each pair of waypoints with a 2 ms spline point spacing.

The algorithm was evaluated for computational performance, trajectory smoothness, 
and the ability of a simulated Kinova Gen3 arm to track the iKinQP generated trajectory with a 
computed torque controller. Our results indicate that the algorithm is capable of running at a 
sub-millisecond cadence, in step with the torque control loop. Generated trajectories are always collision-free, and the produced trajectories were generally smooth enough for a simulated dynamic manipulator to track accurately enough to avoid environmental collisions.

\section{Related Works}

Solving the inverse kinematics problem is traditionally done through analytic or numerical iterative approaches. Analytic approaches work well with six degree-of-freedom robots that admit closed-form solutions, which in practice means that three of the robot's neighboring joint axes intersect to a point \cite{craig2018robotics}. However, when such conditions do not hold, numerical iterative approaches are typically used, which is usually the case for kinematically redundant robots \cite{lynch2017modern}.

% \comment{or on direct numerical search in the joint space
%  - should add something.} 
Classical methods in redundancy resolution for kinematically redundant manipulators 
focus on a Jacobian pseudo-inverse solution \cite{ maciejewski1985obstacle, 
siciliano1990kinematic} or on direct numerical search in the joint space \cite{yao2007path, stilman2010global}. The seminal work on adding collision avoidance considerations to 
Jacobian pseudo-inverse solutions was developed by \citeauthor{maciejewski1985obstacle} 
\cite{maciejewski1985obstacle} in 1985. \citeauthor{maciejewski1985obstacle} based their 
optimization method on the Jacobian pseudo-inverse where a repulsive velocity component 
directs manipulator links away from obstacles when within their safety zones. 
Direct search in the joint space for redundant manipulators subject to task space 
hard constraints fall under the category of constrained motion planning. Such methods use projections onto a 
sub-manifold to ensure end effector constraints are satisfied within an acceptable 
error \cite{ stilman2010global}. However,
classical approaches are limited in where they can be applied due to higher computational costs, complexity, and inefficiency 
at accounting for singularities as well as difficulty with highly curved manifolds in the case of direct search 
methods \cite{kingston2018sampling}. This leads to lower performances than more recent 
approaches like constrained Quadratic Programming (QP).

Quadratic Programming methods have the advantage of generating real-time solutions 
which easily incorporate multiple hard constraints and optimality criteria while also 
performing well near singularities \cite{shankar2015quadratic}. However, many variations 
exist in how collision volumes are treated, and how cost functions are formulated 
\cite{shankar2015quadratic, zhao2018collision, mirrazavi2018unified, zhao2021collision, chembuly2020efficient}.

% When adding in collision avoidance considerations, some methods prioritize trajectory following 
% over collision-free solutions \cite{chembuly2020efficient}. 

% However, such methods would not remain viable 
% if scaled up to environments with multiple robot arms operating in the same space or 
% with humans in the shared workspace. The second strategy instead prioritizes collision 
% avoidance over achieving the planned trajectory to the final position 
% \cite{shankar2015quadratic, zhao2018collision, mirrazavi2018unified, zhao2021collision}. 

\citeauthor{shankar2015quadratic} \cite{shankar2015quadratic} presented an approach using a  
linearly constrained QP to solve the inverse kinematics of redundant manipulators for wheeled, 
legged, or fixed base robots in real-time. It treats collisions using a repulsion method. The approach does not consider preventing high deviations from the current joint position which could possibly result in choppy motions. 

% \textit{\textcolor{blue}{Tie more clearly meaning of these to significance for our paper...}}
\citeauthor{zhao2018collision} \cite{zhao2018collision} simplifies each collision body (manipulator and objects) 
into a sphere or series of overlapping spheres. In doing so this method runs the risk of overfitting the bodies 
and can thus lead to more conservative solutions to the collision avoidance problem. 
Similarly, \citeauthor{mirrazavi2018unified} \cite{mirrazavi2018unified} also 
represent their manipulator model by a series of spheres, but with the opposite 
goal of intercepting target objects rather than avoiding them.
% So
% \cite{mirrazavi2018unified} shares common self-collision constraints with us in their 
% formulation, they also have an opposing constraint of stability upon 
% interception rather than penalizing collisions with external bodies.
 \citeauthor{zhao2021collision} \cite{zhao2021collision} extends 
their collision volume representation to include spheres, cubes, cylinders, etc. allowing 
for more closely representing the different collision bodies in the workspace. However, they 
do not have a self-collision constraint for the arm and necessitate that 
the manipulator be represented by ellipsoids for collision checking. 
\Citeauthor{chembuly2020efficient} \cite{chembuly2020efficient} instead uses a bounding box 
approach to simplify collision checking to bound any obstacle as a box which has a high chance 
of overfitting non-box obstacles. They also treat all collisions as soft constraints.
% They also treat collisions as soft constraints 
% and had much higher computation times than our solution produces.
Our method goes beyond these approaches to allow for more options in how collision bodies are represented, 
giving us the ability to check for a wider range of constraints as hard constraints than these previous methods.

% Some of these hard constraints include collision 
% avoidance (object and manipulator collision) and self-collision (arm 
% colliding with itself) constraints.

Other widely used methods for numerical inverse kinematics solvers 
include Rosen Diankov's Open Robotics 
Automation Virtual Environment (OpenRave)'s IKFast \cite{diankov2008openrave} and Orocos 
Kinematics Dynamics Library (KDL) \footnote{KDL: http://www.orocos.org/wiki/orocos/kdl-wiki/}. 
IKFast is fast but can have infinite solutions in solving some iterations of the inverse kinematics 
problem, as it cannot be used for arms of more than six degrees of freedom. KDL is limited by having 
convergence failures when taking robot joint limits into consideration. A family of KDL open-source extensions 
have been recently developed which do well for redundant 
manipulators in real time -- TRAC-IK \cite{beeson2015trac},  RelaxedIK \cite{rakita2018relaxedik}, CollisionIK \cite{rakita2021collisionik}, and 
RangedIK \cite{wang2023rangedik}. However, the earliest TRAC-IK \cite{beeson2015trac} 
does not consider collisions at all in their formulations. Later open-source versions 
RelaxedIK \cite{rakita2018relaxedik}, CollisionIK \cite{rakita2021collisionik}, and 
RangedIK \cite{wang2023rangedik} add in collision avoidance constraints as soft constraints 
while still ensuring no collisions occur. The main difference between \cite{rakita2018relaxedik, rakita2021collisionik, wang2023rangedik}
and ours is their focus on instantaneous motion generation, whereas we focus on planning full 
trajectories.  

% \added{WE NEED TO ADD MOVEIT!}

% \cite{beeson2015trac, rakita2018relaxedik, rakita2021collisionik, wang2023rangedik} are 
% much slower in generating real-time solutions compared to other popular inverse kinematics 
% solvers in the literature and our proposed solution.
% \textcolor{blue}{\textbf{``much slower" claim may not be correct or the right way of phrasing this...
% the main difference between RelaxedIK and CollisionIK with iKinQP may be that they are focused 
% on instantaneous planning while ours is focused on trajectory planning}}

% \citeauthor{chembuly2020efficient} \cite{chembuly2020efficient} devised a different formulation to the 
% optimization problem for solving the inverse kinematics trajectory 
% planning problem. Their formulation highly penalizes collisions in the minimization 
% function rather than treating collision as a constraint. Further, \citeauthor{chembuly2020efficient}
% devise their hard constraints differently as they constrain the problem
% to encourage the end effector reaching a target position. \textit{\textcolor{blue}{Tie 
% this back to our work/why this being formulated differently would matter or be 
% something worth mentioning...}}

Another set of approaches for full trajectory planning with collision avoidance for redundant 
manipulators comes from motion planning \cite{gasparetto2015path, kingston2018sampling}. 
A common strategy in motion planning is to combine an off-line path 
planning algorithm and on-line motion control.
Path planning acts as a trajectory generator which plans to avoid 
fixed obstacles known a priori to the robot. On-line motion 
control adds in a way to deal with dynamic obstacles and avoid
singularities \cite{palmieri2021motion}. However, while traditional motion 
planning algorithms tend to do very well in finding asymptotically optimal solutions 
\cite{karaman2011sampling} they typically take a long time to converge to such solutions. Some newer 
methods incorporate task and motion planning to deal with symbolic sequencing in 
addition to geometric constraints, but still base trajectory planning of traditional motion planners and can have 
similar limitations \cite{ren2021extended}. Other newer methods aim to combine 
motion planning with trajectory optimization for shortest path trajectory planning, using safety 
sets to completely avoid collisions \cite{marcucci2022motion}. This differs from our proposed method as they 
instruct the robot to stay within certain pre-defined safe regions while we focus
on defining unsafe collision objects to avoid.

The popular motion planning and mobile manipulation framework MoveIt! 
\cite{chitta2012moveit, sucan2019moveit} is robot-agnostic and works for one or two arm cases. 
Its motion planning capabilities are based on OMPL 
\cite{sucan2012the-open-motion-planning-library}, a motion planning library containing 
state of the art sampling-based motion planning algorithms under the categories of geometric, 
control-based, and multilevel-based planners. However, MoveIt! has a few notable limitations. 
MoveIt! generates free motions, meaning there is no control over the end effector’s path to the 
goal, and creates randomized plans where repeatability is not guaranteed. Further, MoveIt! does 
not account for any velocity control of the end effector's motion. \cite{fresnillo2023extending} 

% are not as light weight, smooth, or fast as our proposed solution. \textbf{Last 
% sentence claim is very strong would need to back it up or soften it!!}
% \textbf{\textcolor{red}{Back 
% this up/make sure we prove this claim!}}

Data-driven methods for solving inverse kinematics problems are growing in popularity 
as methods for dealing with redundant kinematics and constrained environments \cite{hassan2020inverse}.
A recent example, IKFlow, \cite{ames2022ikflow} generates a high number of solutions for poses for redundant manipulators following 
a trajectory. However, IKFlow \cite{ames2022ikflow} necessitates having a one-off training period to generate solutions 
for each new robot used with the system.
% and is slower with much higher errors than iKinQP. 
% \textbf{\textcolor{red}{Back 
% this up/make sure we prove this claim!}}/ 
This represents the drawback of learning methods in 
trajectory planning problems as such methods tend to require large datasets, initial training time, 
and cannot as easily adapt to clutter. 

% \comment{This isn't true of RRT methods, 
% which we absolutely need to address here. MoveIT! is the best-known example.}

% There are a whole bunch of strategies that can do obstacle 
% avoidance using all sorts of different sensor modalities, 
% but one nifty thing about iKinQP is that (with some prior 
% knowledge of the environment) you don’t really need real 
% time sensing other than robot joint sensors. ... (cite 
% something to back this??) \
% \textit{\textcolor{blue}{Should I try to include a statement like this here...?}}

\section{iKinQP Algorithm}

The six-spatial-degree-of-freedom robot Jacobian $J(\mathbf{q})$ in (\ref{eq:jacobian}) provides a kinematic mapping between joint space velocities, $\dot{\mathbf{q}}$, and the corresponding Cartesian translational velocities, $\dot{\mathbf{p}}$, as wells as the rotational velocities, $\dot{\boldsymbol{\omega}}$, of a point rigidly fixed to the manipulator given joint orientation, $\mathbf{q}$.

\begin{equation}
\begin{pmatrix}
\dot{\mathbf{p}} \\
\dot{\boldsymbol{\omega}} 
\end{pmatrix} 
= \mathbf{v}
= J(\mathbf{q})\dot{\mathbf{q}}
\label{eq:jacobian}
\end{equation}

Given (\ref{eq:jacobian}), the inverse kinematics problem of finding a suitable joint space velocity trajectory, $\dot{\mathbf{q}}_d$, to follow a desired Cartesian space velocity trajectory, $\mathbf{v}_{d}$, can be stated as follows.

\begin{equation}
    \min_{\dot{\mathbf{q}}_d} ||\mathbf{v}_{d} - J(\mathbf{q})\dot{\mathbf{q}}_d ||   \label{eq:vel_err}
\end{equation}

We can keep the newly computed solution close to the current joint configuration and minimize integrator drift by adding the following term to be minimized:

\begin{equation}
    ||\dot{\mathbf{q}}_{d} - (\mathbf{q}+\delta t *\dot{\mathbf{q}}) ||   \label{eq:pos_err}
\end{equation}

However, when dealing with kinematically redundant manipulators that have seven or more actuated degrees-of-freedom, (\ref{eq:vel_err}) becomes an under-constrained problem since there are only six spatial degrees-of-freedom (three translational and three rotational). The problem can become fully constrained by adding additional constraints that make use of the redundant actuated degree(s) of freedom to improve the quality of the selected trajectory. These can be either ``soft'' constraints which are appended to the minimization term above, or ``hard'' constraints that are enforced by the optimizer. For instance, a soft constraint that keeps joint rates as small as possible and increases trajectory smoothness is:

\begin{equation}
    ||\dot{\mathbf{q}}_{d} ||   \label{eq:vel_tot}
\end{equation}

We can also add hard constraints that reflect conditions the robot arm cannot be allowed to violate, such as joint limits and joint velocity limits:

\begin{equation}
\begin{split} 
    \min_{\dot{\mathbf{q}}_d}&( ||\mathbf{v}_{d} - J(\mathbf{q})\dot{\mathbf{q}}_d ||  + \gamma||\dot{\mathbf{q}}_{d} - (\mathbf{q}+\delta t *\dot{\mathbf{q}}) || + \lambda||\dot{\mathbf{q}}_{d} ||) \\
    s.t.& \; \; q_{i,lb} \leq q_{i,d} \leq q_{i,ub} \\
    & \; \; \dot{q}_{i,lb} \leq \dot{q}_{i,d} \leq \dot{q}_{i,ub}
\end{split}
\label{eq:cost}
\end{equation}

where the constraints in (\ref{eq:cost}) are for each desired joint position $q_{i,d}$ and each desired joint velocity $\dot{q}_{i,d}$, where $i \in [1,N]$ for a manipulator with N actuated degrees-of-freedom. $\lambda$ and $\gamma$ are constant scalars or matrices that magnify the relative importance of their respective terms. 

To enforce collision avoidance, we can add a constraint that requires all parts of the robot to maintain a minimal buffer distance, $d_{\text{buff}}$, from all other parts of the robot, $d_{robot,robot}$ and from other environmental objects, $d_{robot,env}$:

\begin{equation}
    \begin{split} 
        \min([d_{robot,robot}(\mathbf{q}_d),d_{robot,env}(\mathbf{q}_d)]) \geq d_{\text{buff}}
    \end{split}
\label{eq:collision1}
\end{equation}

For this paper, collisions objects are represented as kinematic trees of elementary collision volumes. The single file collision detection library NTCD\cite{Grieverheart2017} is used to compute the shortest distance between each pair of elementary geometries. 

It is important to note that in (\ref{eq:cost}) and (\ref{eq:collision1}), joint limits and collisions are treated as hard constraints, whereas Cartesian tracking error is a soft constraint. This allows for the computed trajectory to deviate from the desired in order to strictly avoid joint limits and collisions. 

The formulation provided above cannot immediately be implemented, as the joint constraints are always aligned, and constrained optimization solvers generally require the problem to be expressed in ``normal'' form. This work utilizes qpOASES\cite{Ferreau2014} to solve the constrained optimization problem. qpOASES requires problems of the following form:

\begin{equation}
\begin{split} 
    \min_{a} \; \; &\frac{1}{2}\mathbf(a)^{\intercal}H\mathbf{a} + \mathbf{a}^{\intercal}g(w_0)  \\ 
    s.t. \; \; & lbA(w_0) \leq A\mathbf{a} \leq ubA(w_0) \\
    & lb(w_0) \leq \mathbf{a} \leq ub(w_0)
\end{split}
\label{eq:QPoases}
\end{equation}

In (\ref{eq:QPoases}), $\mathbf{a}$ is the generic variable being optimized. For this work the variable being optimized is the joint velocity trajectory, $\dot{\mathbf{q}}$, which can be integrated to get the corresponding joint position trajectory, $\mathbf{q}$. Rearranging the expression to be minimized from (\ref{eq:cost}), $H$ and $g$ can be expressed as follows.

\begin{equation}
    \begin{split} 
        H &= J^{\intercal}(\mathbf{q})J(\mathbf{q}) + \delta t^2 * J^{\intercal}(\mathbf{q})*\gamma * J(\mathbf{q}) + \lambda I \\ 
        g &= -J^{\intercal}(\mathbf{q})\dot{\mathbf{x}}_d + \delta t * J^{\intercal}(\mathbf{q})*\gamma * (\mathbf{x}-\mathbf{x}_d)
    \end{split}
  \label{eq:hessian}
\end{equation}

In (\ref{eq:hessian}), $\mathbf{x}_d$, is a vector representing the desired Cartesian position and orientation, and $\mathbf{x}$ is the current Cartesian position and orientation computed by performing forward kinematics on the current joint configuration $\mathbf{q}$. We note here that the orientation component of $\mathbf{x}$ and $\mathbf{x}_d$ can be represented as desired, however even though all representations express three degrees-of-freedom, conventions including quaternions and axis-angle require four terms. For the purposes of this paper, forward kinematics and kinematic terms including the Jacobian $J$ are solved for using the open-source Pinocchio\cite{pinocchioweb} rigid body dynamics library.

In order to resolve the fact that the joint constraints are aligned in the initial formulation, the two joint constraint inequalities can be combined to bound $\dot{\mathbf{q}}_d$.

\begin{equation}
    \begin{split} 
        lb &= \max ([\dot{\mathbf{q}}_{\mathit{low\ limit}},(\mathbf{q}_{\mathit{low\ limit}}-\mathbf{q})/ \delta t]) \\ 
        ub &= \min ([\dot{\mathbf{q}}_{\mathit{high\ limit}},(\mathbf{q}_{\mathit{high\ limit}}-\mathbf{q})/ \delta t]) 
    \end{split}
 \label{limits}
\end{equation}

Finally, the collision avoidance constraint in (\ref{eq:collision1}) is expressed in terms of $\mathbf{q}_d$ rather than a linear function of the optimization variable $\dot{\mathbf{q}}_d$ as required. Note that the first order Taylor expansion of the distance between two collision trees, $d_{tree1,tree2}(\mathbf{q})$ about the current orientation $\mathbf{q}$ perturbed by some small $\boldsymbol\epsilon$ is:

\begin{equation}
        d_{\mathit{tree1,tree2}}(\mathbf{q+\boldsymbol\epsilon})\approx d_{\mathit{tree1,tree2}}(\mathbf{q}) + \frac{ \partial d_{\mathit{tree1,tree2}}(\mathbf{q})}{ \partial q}\boldsymbol\epsilon    
 \label{eq:taylor}
\end{equation}

Assuming a constant joint velocity trajectory, $\dot{\mathbf{q}}_d$, over the perturbation duration $\delta t$, $\boldsymbol\epsilon$ can be replaced with $\dot{\mathbf{q}}_d\delta t$ and (\ref{eq:taylor}) becomes:

\begin{equation}
    d_{\mathit{tree1,tree2}}(\mathbf{q+\boldsymbol\epsilon})\approx d_{tree1,tree2}(\mathbf{q}) + \frac{ \partial d_{\mathit{tree1,tree2}}(\mathbf{q})}{ \partial \mathbf{q}}\dot{\mathbf{q}}_d\delta t    
\label{eq:taylor2}
\end{equation}

The collision avoidance constraint can now be formulated as follows.
\begin{equation}
    \frac{ \partial d_{tree1,tree2}(\mathbf{q})}{ \partial \mathbf{q}}\delta t \dot{\mathbf{q}}_d \geq d_{\text{buff}}  
\label{eq:collision2}
\end{equation}

From (\ref{eq:collision2}) the parameters of the first set of constraints in (\ref{eq:QPoases}) can be defined. $A=\frac{ \partial d_{tree1,tree2}(\mathbf{q})}{ \partial \mathbf{q}}\delta t$, $lbA = d_{\text{buff}}$, and $ubA$ is selected to be some arbitrarily large value.

It is difficult to find a general analytic solution to the gradient of the distance function. Instead, the gradient is solved for numerically via symmetric difference quotient at each joint index $i$ by perturbing each joint angle $q_i$ by $\delta q$, and leaving the other joint values unchanged. $\delta q$ is chosen to be much smaller than the normal change in joint angles over a time step $\delta t$. In (\ref{eq:gradient}) below, $\delta \mathbf{q}_i$ is a vector the same size as $\mathbf{q}$ with all components set to zero except for the $i$th component which is set to $\delta q$.

\begin{equation}
    \begin{split}
    &\frac{ \partial d_{\mathit{tree1,tree2}}(\mathbf{q})}{ \partial \mathbf{q}}[i] \approx\\
    & \frac{d_{\mathit{tree1,tree2}}(\mathbf{q + \delta \mathbf{q}_i }) - d_{\mathit{tree1,tree2}}(\mathbf{q_i - \delta \mathbf{q}_i })}{2\delta q}
    \end{split}
\label{eq:gradient}
\end{equation}

\section{Experimental Setup}

In order to evaluate our algorithm we created three simulation test scenarios using MuJoCo \cite{todorov2012mujoco} Version 2.1.0. For each of these scenarios, one or more robotic arms were provided a series of task-space waypoints in the world frame for the end-effector(s) to achieve. The positions of these waypoints were randomly selected along the surface of an imaginary 0.9 m radius sphere, and the corresponding orientations were computed to be perpendicular to the sphere's surface. Waypoint terminal velocities were always set to zero for these tests, although the algorithm is flexible enough to allow arbitrary terminal velocities. A versine-ramp function \cite{yavuz2012hybrid} was used to interpolate between the current end-effector position and the upcoming waypoint over a predefined trajectory duration, $T_{traj}$. 

\begin{figure}[htbp]
    \includegraphics[scale=0.38]{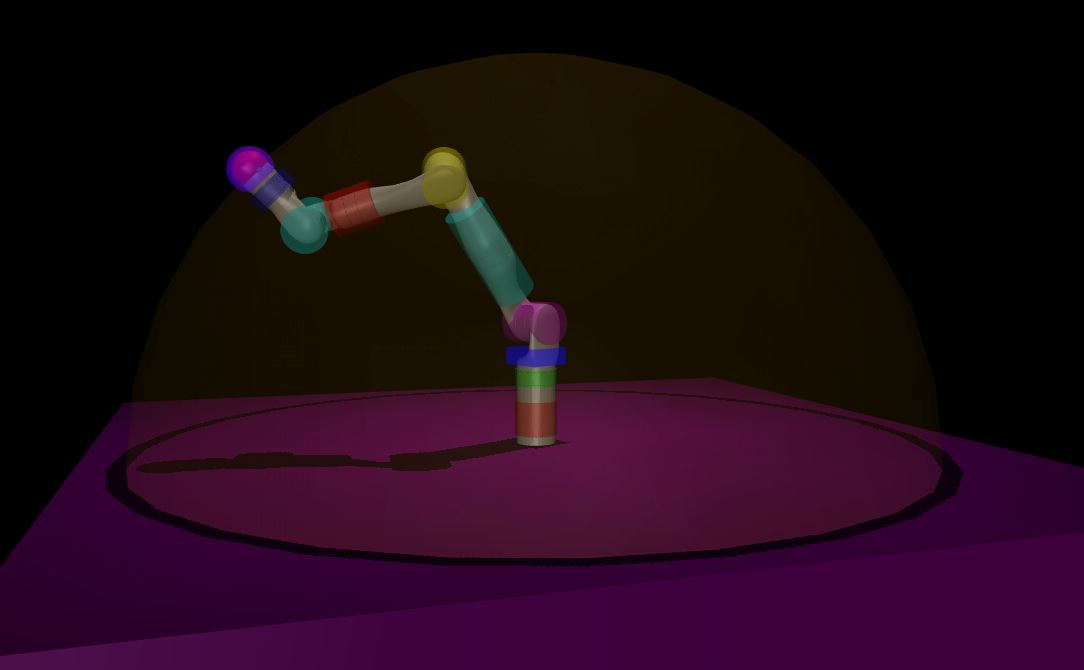}
    \caption{MuJoCo simulation of Kinova arm (gray) overlaid with multicolor collision volumes. Environmental collision volume is the floor (magenta). Random task-space waypoints are provided to the arm with positions along a sphere (orange) of 0.9 m radius, and orientations perpendicular to the sphere.}
    \label{fig:1armfloor}
\end{figure}

In the first scenario, a single Kinova Gen3 arm was mounted to the ground as in Fig. \ref{fig:1armfloor}. This task demonstrates the ability of iKinQP to produce a smooth joint-space trajectory between task-space waypoints that avoids joint position and velocity limits, avoids arm self-collisions, and avoids collisions between the arm and the ground. This is a mostly ideal scenario since, other than the arm itself, there are not any obstacles in the workspace.

\begin{figure}[htbp]
    \includegraphics[scale=0.40]{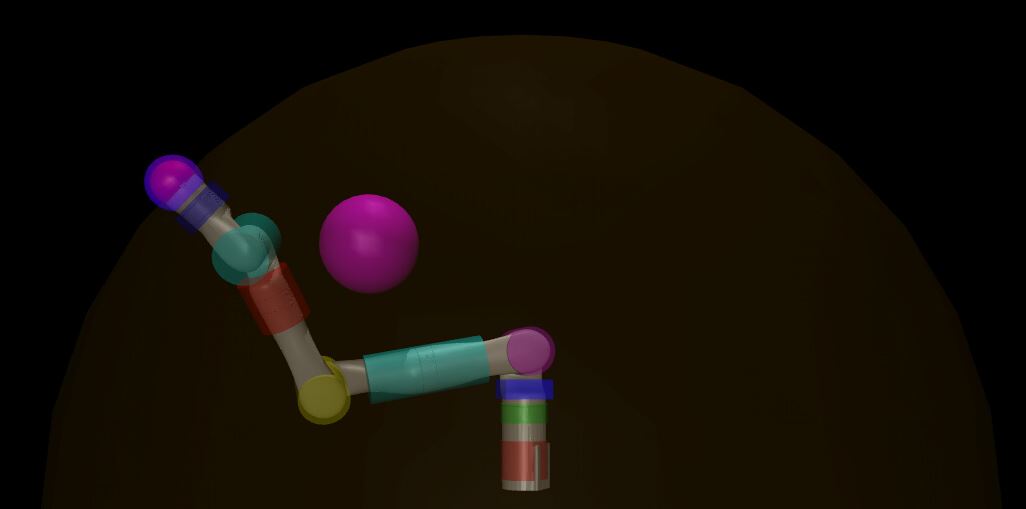}
    \caption{MuJoCo simulation of Kinova arm (gray) overlaid with multicolor collision volumes. An environmental collision volume (magenta sphere) is placed in the workspace. Random task-space waypoints are provided to the arm with positions along a sphere (orange) of 0.9 m radius, and orientations perpendicular to the sphere.}
    \label{fig:1armsphere}
\end{figure}

In the second scenario, a single Kinova Gen3 arm was rigidly fixed to the environment, but for this scenario there was a large spherical collision volume obstructing a significant portion of the workspace as in Fig. \ref{fig:1armsphere}. This task demonstrates the ability of iKinQP to produce smooth collision-free joint-space trajectories, subject to joint limits, even when there is a stationary object obstructing a portion of the workspace.

\begin{figure}[htbp]
    \includegraphics[scale=0.44]{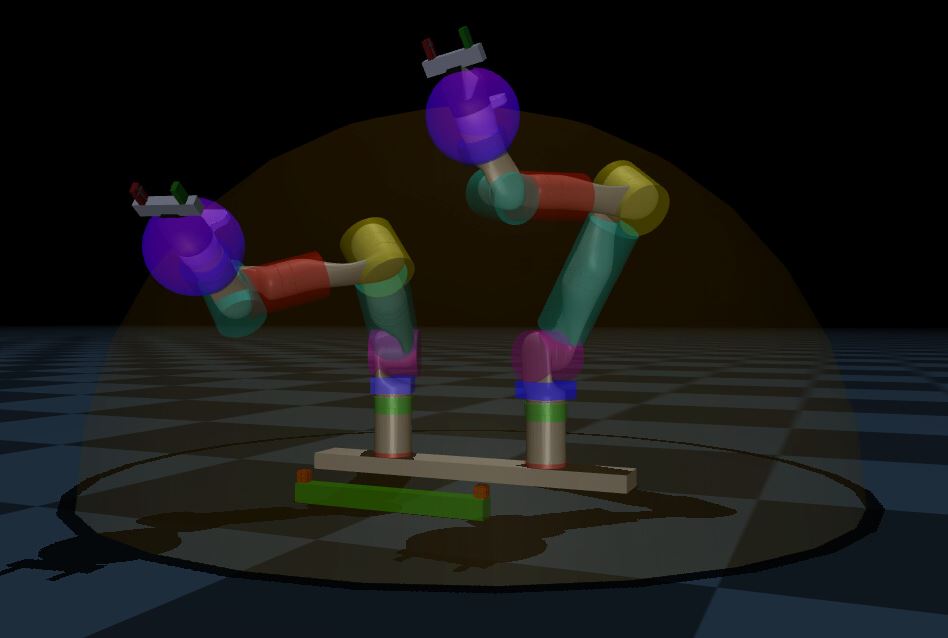}
    \caption{MuJoCo simulation of two Kinova arms (gray) overlaid with multicolor collision volumes. The 2 arms are rigidly mounted to a common structure. Random task-space waypoints are provided to each arm with positions along a sphere (orange) of 0.9 m radius, and orientations perpendicular to the sphere. The two arms must figure out how to reach their respective waypoints while avoiding each other.}
    \label{fig:2arm}
\end{figure}

\begin{figure}[htbp]
    \centering
    \includegraphics[scale=0.5]{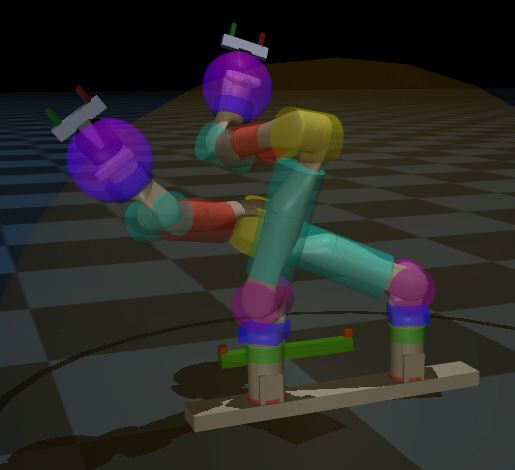}
    \caption{Two arms are working to achieve their target waypoints while avoiding each other.}
    \label{fig:tangled}
\end{figure}

For the third scenario a pair of Kinova Gen3 arms were rigidly mounted to a beam, and the arms were separated by a distance of 0.4 m as in Fig. \ref{fig:2arm}. The task-space waypoints were produced on a spherical surface with radius 0.9 m centered between the two arms. Each arm was provided its own task-space waypoint along the surface of the sphere. iKinQP was used to coordinate smooth joint-space trajectories such that the two arms did not collide with each other, even if the arms needed to crisscross with each other to achieve their desired waypoint as in Fig. \ref{fig:tangled}. 

\section{Results and Discussion}

We evaluated the performance of iKinQP regarding three categories for each of the scenarios described in the previous section. 

First we evaluated the computational performance of the algorithm. We computed the median CPU clock duration for iKinQP to settle on a solution for each control step. This is the total amount of CPU time required to compute a reference joint space position and velocity trajectory from a Cartesian reference position and velocity. A median rather than a mean was used here since computations were not performed on a real-time operating system, and system noise can skew the mean. Then we report the mean number of working set recomputes (nWSR), and the number of active constraints (NAC) required for each solution.

Second we computed the mean tracking error of the joint-velocity profile produced by iKinQP. While it is guaranteed that the trajectory produced by iKinQP is collision-free, it is not guaranteed that the dynamics of the robot allow the arm to perfectly track this trajectory. This is because iKinQP is formulated using kinematics in the absence of knowledge of robot dynamics. Neglecting dynamics is one of the main contributors to the computational efficiency of iKinQP. 

Finally, we evaluated the "smoothness" of the trajectories for each test case. It is expected that trajectory smoothness is directly correlated with how well the robotic arm is able to track the iKinQP produced joint trajectory. For this paper, we assess smoothness by providing a histogram of the joint trajectory jerk in the time domain, as well as a Fourier transform of the joint position trajectory. Both of these are given for the most challenging two-arm test case. 

For the test scenarios described in the previous section we evaluate the results for when the arm is requested to achieve the requested waypoint in 15 seconds, and for a more aggressive scenario when the arm is requested to achieve the requested waypoint in 5 seconds causing joint velocity limits to come into play.

\subsection{Computational Time}

\begin{table}[htbp]
    \caption{Computational Performance Values. $T_{traj}$: Allotted trajectory duration to get from current Cartesian coordinates to desired Cartesian waypoint. nWSR: Number of Working Set Recomputes. NAC: Number of Active Constraints. Scenario 1: A single arm avoiding a floor collision volume. Scenario 2: A single arm avoiding a sphere collision volume. Scenario 3: Two arms avoiding each other.}
    \begin{center}
        \begin{tabular}{ | l | l | l | l | l | }
        \hline
        \textbf{Scenario} & \parbox[ct]{1.0cm}{ $T_{\mathit{traj}}$ (s) } & \parbox[ct]{1.0cm}{\textbf{Median CPU Time} (ms) } & \parbox[ct]{1.0cm}{ \textbf{Mean nWSR}} & \parbox[ct]{1.0cm}{ \textbf{Mean NAC}} \\ \hline
        1 & 15&  0.26& 10.15& 0.02\\ \hline
        1 &  5&  0.28&  9.78& 0.06\\ \hline
        2 & 15&  0.29&  9.60& 0.17\\ \hline
        2 &  5&  0.26&  9.98& 0.18\\ \hline
        3 & 15&  1.47& 23.06& 1.59\\ \hline
        3 &  5&  1.64& 24.73& 1.98\\ \hline
        \end{tabular}
    \end{center}
\label{tab:comp_time}
\end{table}

The iKinQP algorithm was written in C++, and run on an Intel(R) Core(TM) i9-10900 2.80GHz CPU with 64 GB RAM. The code was not multithreaded. Nonetheless, computational performance was quite good and very reasonable for real-time robotic control in step with torque level control.  

Table \ref{tab:comp_time} shows that the median computational time for all one arm tests was less than 0.3 ms, with standard deviations between 3.3 and 4.6 ms. Each solution required on average about 10 working set recomputes (nWSR). Constraints were rarely active during the tests where the arm needed to avoid the floor, and about three times more frequent for the test where the arm needed to avoid the spherical collision volume in the middle of the workspace. For this later scenario a significant increase in nWSR likely wasn't observed because active collisions were still overall fairly infrequent. 

\begin{figure}[htbp]
    \includegraphics[scale=0.6]{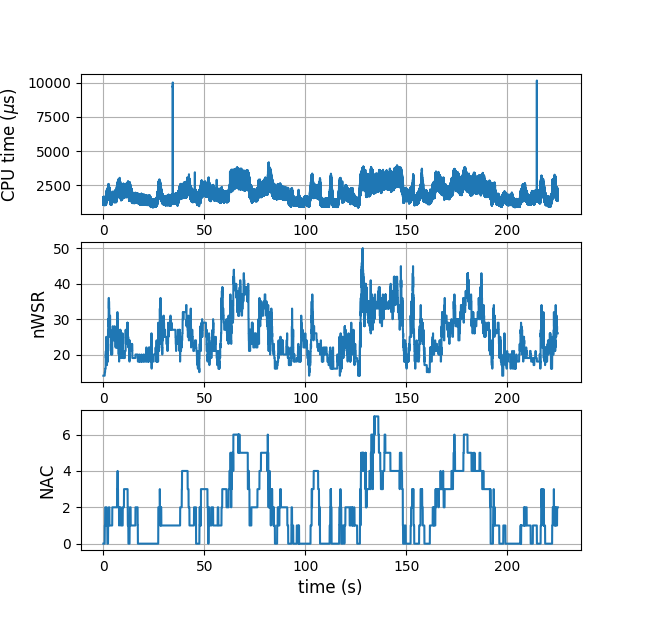}
    \caption{Computational performance for the most challenging test case (two arms avoiding each other with waypoint spacing of 5 seconds). NAC drives performance. nWSR: Number Working Set Recomputes; NAC: Number Active Constraints.}
    \label{fig:2arm_time_hist}
\end{figure}

For the two arm tests, median computation time was 1.47 s for the more relaxed $T_{\mathit{traj}}=15$ s test, and 1.64 for the more aggressive $T_{\mathit{traj}}=5$ s test. The increase in compute time for this more aggressive test appears to correlate with the increased number of average active constraints, and the average number of working set recomputes as shown in Fig. \ref{fig:2arm_time_hist}. For the two arm tests, the size of the self-collision tree that needed to be searched each iteration was doubled, therefore a significant slow down is expected with the addition of each successive arm. Nonetheless, even for this scenario  iKinQP iterations most frequently took less than 1ms. For experiments requiring more than two arms, code parallelization and optimization might be required to maintain real-time rates depending on the application. 

\subsection{Tracking performance}

While the iKinQP produced trajectories are guaranteed to avoid collisions, it cannot be guaranteed that a robot following these trajectories will avoid collisions unless the robot is tracking the trajectories perfectly. For this reason, we provide results for the Kinova arm tracking performance when the joint trajectory references from iKinQP were passed to a computed torque controller. The tracking position error was computed at each time for each joint as: 

\begin{equation}
    \tilde{\mathbf{q}} = \mathbf{q}_{ref} - \mathbf{q}_{\mathit{meas}}
\label{track err}
\end{equation}

The tracking metric reported in Table \ref{tab:track_err} is the absolute value of the mean of the tracking error plus one standard deviations for all joints, reported in mrad.  

\begin{table}[htbp]
\caption{Joint Position Tracking Error, mean plus standard deviation in mrad. $T_{traj}$: Allotted trajectory duration to get from current Cartesian coordinates to desired Cartesian waypoint. nWSR: Number of Working Set Recomputes. NAC: Number of Active Constraints. Scenario 1: A single arm avoiding a floor collision volume. Scenario 2: A single arm avoiding a sphere collision volume. Scenario 3: Two arms avoiding each other.}    
\begin{center}
    \begin{tabular}{ | l | l | l | }
    \hline
    \textbf{Scenario} & \parbox[ct]{1cm}{ $T_{\mathit{traj}}$\\ (s) } & \parbox[ct]{2cm}{$|\boldsymbol{\mu}(\tilde{\mathbf{q}})| + \boldsymbol{\sigma}(\tilde{\mathbf{q}})$\\(mrad)} \\ \hline
    1 & 15 & 2.45 \\ \hline
    1 & 5  & 5.75 \\ \hline
    2 & 15 & 3.31 \\ \hline
    2 & 5  & 5.81 \\ \hline
    3 & 15 & 6.55 \\ \hline
    3 & 5  &11.65 \\ \hline
    \end{tabular}
\end{center}
\label{tab:track_err}
\end{table}

Table \ref{tab:track_err} shows that overall tracking performance was quite reasonable, with the magnitude of tracking errors primarily in the single digits of milliradians or less. We would expect that when given a longer time to achieve a waypoint, the tracking error would be smaller. The results agree with this hypothesis showing that a 15-second trajectory duration resulted in smaller errors. Also as expected, slightly more error was observed for the scenario shown in Fig \ref{fig:1armsphere} (where iKinQP produced a trajectory that avoided a sphere in the middle of the workspace) than the scenario shown in Fig. \ref{fig:1armfloor} (where iKinQP produced a trajectory that avoided the floor). This is likely because collisions are hard constraints. When there are active constraints from collision volumes in close-proximity, optimization of the cost function (including minimizing trajectory jerk) is treated as a secondary task of the QP optimizer. 

The scenario shown in Fig. \ref{fig:2arm} was the most challenging scenario and correspondingly resulted in the largest tracking errors. It is supposed that in this scenario, all the hard constraints imposed by the various moving collision volumes and joint limits disqualify a significant portion of the configuration space. Maneuvering between the valid portions of the configuration space, which are constantly changing, creates a trajectory which is less smooth, and therefore harder for a real dynamical system to track.

%TODO: Make sure to plot joint limits when plotting joint velocity trjactory

\begin{figure}[htbp]
    \includegraphics[scale=0.6]{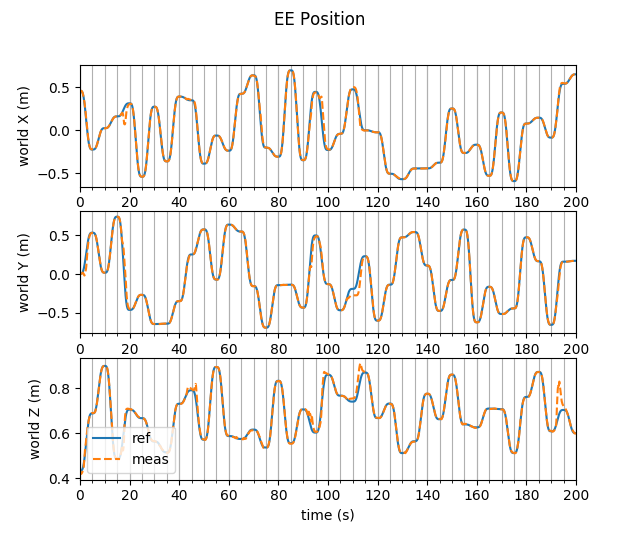}
    \caption{Ability of robot arm to follow desired Cartesian end-effector trajectory. The results depicted are for the most challenging one-arm scenario where waypoints are spaced by 5s and a sphere collision volume is in the middle of the workspace. Observed differences between the reference trajectory and measured trajectory are primarily due to a shift in the (interim) reference trajectory by iKinQP to avoid collisions, such as around 105s and 195s.}
    \label{fig:1arm_sphere_avoid_time}
\end{figure}

The above analysis did not mention task-space tracking performance. This is because while we generally want the robot to achieve perfect task-space tracking, we do not want the robot to track the task-space trajectory when this would cause a collision. Fig. \ref{fig:1arm_sphere_avoid_time} compares the reference Cartesian position trajectory with what was actually observed for the single arm test where an 0.1 m radius sphere was placed in the workspace at a position of (-0.25 m, 0.2 m, 0.5 m). In the figure it is apparent that the arm achieved near perfect tracking of the trajectory except when it was near the spherical collision volume. This is exactly the desired behavior.

\subsection{Trajectory Smoothness}

To evaluate trajectory smoothness first the jerk components of the joint trajectory are analyzed using a histogram. As a reference for understanding the histogram results, we created two test profiles: one that we expected to be smooth, and one that we expected to be rough. The smooth profile was created by generating a sinusoid with a period of 5 seconds (the shortest trajectory duration we tested with) and an amplitude of 1.39 rad/s (the joint velocity limit of the Kinova arm). This sinusoid was then numerically double differentiated over the sampling time step of 2 ms using the numpy gradient function. The rough profile was generated by pulling numbers between $\pm 1.39$ from a zero-mean random-uniform distribution. This profile was also numerically double differentiated over the sampling time step of 2 ms using the numpy gradient function.

\begin{figure}[htbp]
    \includegraphics[scale=0.6]{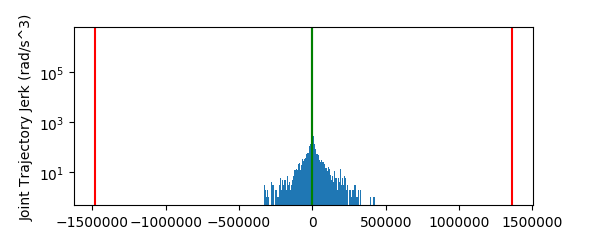}
    \caption{Histogram of joint trajectory jerk for all the 14 joints during a two arm test with 5 s trajectory durations (the most challenging test case). Green vertical lines show the maximum and minimum jerk values of a sinusoid velocity trajectory. Red vertical lines show the maximum and minimum jerk values when velocities are chosen from a uniform random distribution. }
    \label{fig:2arm_jerk_hist}
\end{figure}

Fig. \ref{fig:2arm_jerk_hist} shows a 1000 bin histogram of the raw jerk values for all fourteen joints during the two arm self-collision avoidance test with a trajectory duration of 5 s. Even for this most aggressive test, jerk values near 0 $rad/s^3$ were much more common than other jerk values. A log scale was required in the y-axis in order to see any of the other bins. Figure \ref{fig:2arm_jerk_hist} also shows that the spread of jerk values for this most aggressive test is significantly wider than for the reference smooth sinusoid profile, but significantly narrower than for the reference rough random trajectory. 

\begin{figure}[htbp]
    \includegraphics[scale=0.6]{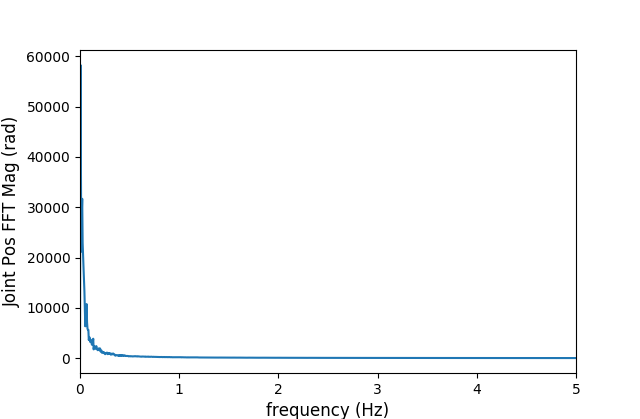}
    \caption{Magnitude of the fast-Fourier-transform (FFT) of the iKinQP produced joint position trajectory for the two arm test with 5 s trajectory durations (the most challenging case). The DC component is removed from the signal before taking the FFT, and the displayed magnitude is magnitude averaged over the 14 joints. Results flat-line at about 0.5 Hz and therefore higher frequency magnitudes are not displayed. }
    \label{fig:2arm_pos_traj_hist}
\end{figure}

For the same two-arm scenario, the joint angle trajectory produced by iKinQP was also evaluated in the frequency domain using a fast-Fourier-transform (FFT), as shown in Fig. \ref{fig:2arm_pos_traj_hist}. The DC component of the signal was removed before computing the FFT, and magnitude values for each frequency bin were averaged over the 14 joints. Results for frequency bins greater than 5 Hz were not displayed as they are all close to zero. From the figure it is observed that the joint trajectory frequency content primarily resides in the 0 to 0.5 Hz range, and there are no obvious spikes at particular frequencies. This is indicative of a smooth trajectory that should be well within the control bandwidth of the system. Precaution can be taken to notch-filter out particular ultra-low-frequency content that could potentially interact with structures that the manipulators are mounted to. 

\section{Conclusion} 
We have presented iKinQP: a quadratic programming based algorithm for generating smooth, real-time, collision-free, joint-space trajectories between task-space waypoints while adhering to joint limits. We demonstrated the performance of our algorithm for a single arm presented with static environmental collision volumes, and for a pair of two 7-dof Kinova Gen3 arms dynamically avoiding each other. Computational performance of the algorithm, trajectory smoothness, and the ability of the dynamical robotic system to track the trajectory were evaluated. Results confirm that iKinQP produces smooth, collision-free trajectories at real-time clock rates in step with torque control for robotic applications. 

\section*{Acknowledgments} 
This research was developed with funding from the Defense Advanced Research Projects Agency (DARPA) in support of the Robotic Servicing of Geosynchronous Satellites (RSGS) program. The views, opinions, and/or findings expressed are those of the author(s) and should not be interpreted as representing the official views or policies of the Department of Defense or the U.S. Government.
\bibliographystyle{unsrtnat} %unsrt
\bibliography{references.bib} 

\end{document}